# CCSRP: Robust Pruning of Spiking Neural Networks through Cooperative Coevolution

Zichen Song, Jiakang Li, Songning Lai, Sitan Huang
(Co-First Author)

**Abstract.** Spiking neural networks (SNNs) have shown promise in various dynamic visual tasks, yet those ready for practical deployment often lack the compactness and robustness essential in resource-limited and safety-critical settings. Prior research has predominantly concentrated on enhancing the compactness or robustness of artificial neural networks through strategies like network pruning and adversarial training, with little exploration into similar methodologies for SNNs. Robust pruning of SNNs aims to reduce computational overhead while preserving both accuracy and robustness. Current robust pruning approaches generally necessitate expert knowledge and iterative experimentation to establish suitable pruning criteria or auxiliary modules, thus constraining their broader application. Concurrently, evolutionary algorithms (EAs) have been employed to automate the pruning of artificial neural networks, delivering remarkable outcomes yet overlooking the aspect of robustness. In this work, we propose CCSRP, an innovative robust pruning method for SNNs, underpinned by cooperative co-evolution. Robust pruning is articulated as a tri-objective optimization challenge, striving to balance accuracy, robustness, and compactness concurrently, resolved through a cooperative co-evolutionary pruning framework that independently prunes filters across layers using EAs. Our experiments on CIFAR-10 and SVHN demonstrate that CCSRP can match or exceed the performance of the latest methodologies.

## 1   Introduction

In recent years, spiking neural networks (SNNs) have achieved significant success in the field of dynamic vision, such as spiking neural state image classification and object detection. Despite their impressive performance, the high computational cost associated with converting artificial neural networks (ANNs) to SNNs limits their deployment in resource-constrained scenarios.[1] Additionally, SNNs are susceptible to malicious attacks, posing a challenge to their reliability in safety-critical environments. Thus, enhancing both the compactness and robustness of SNNs is crucial in many practical applications, such as event cameras.[2]

However, most previous work has focused solely on enhancing either the compactness or robustness of spiking neural networks.[3] On one hand, various model compression techniques have been proposed to reduce the computational cost of SNNs, such as neural network pruning and quantization. Among these, neural network pruning aims to remove redundant parameters in networks while maintaining accuracy, and has achieved considerable success. On the other hand, methods like adversarial training, which aim to minimize training loss on adversarial examples, can significantly enhance the robustness of SNNs.[4]

Recent efforts have considered network robustness in the context of pruning SNNs. Typically, these approaches use expert-designed criteria to measure the importance of network weights and prune accordingly. However, designing and tuning such criteria require extensive expertise and laborious experimentation, making them difficult to apply in practical scenarios with diverse datasets and SNN architectures. Moreover, these efforts mainly focus on unstructured neural network pruning, which hardly reduces computational costs in real-world applications due to the resulting irregular structures being incompatible with mainstream software and hardware frameworks. [5] Thus, an automated structured robust pruning method is essential for practical applications. Robust pruning of SNNs can naturally be framed as an optimization problem, aiming to find a subnet of the original network that maintains high accuracy and robustness but with lower computational cost.[6] Evolutionary algorithms (EAs), inspired by natural evolution, have been used for automatically pruning SNNs. However, unlike artificial neural networks from the last century, modern SNNs typically comprise dozens of layers and millions of parameters, implying a vast search space. For EAs, finding satisfactory solutions within a limited computational overhead is challenging. Recently, Shang et al. proposed an evolutionary pruning method inspired by cooperative co-evolution, CCEP, which has shown encouraging results for large-scale pruning problems. However, their focus was solely on accuracy without considering robustness. In this paper, we introduce a novel Cooperative Coevolutionary Strategy for Robust Pruning (CCSRP).[7] The robust pruning problem is explicitly formulated as a three-objective optimization problem, aiming to simultaneously optimize accuracy, robustness, and compactness. A cooperative co-evolution framework is employed to tackle the robust pruning problem, dividing the search space by layer and applying an EA to optimize each group independently. Additionally, to address the time-consuming process of generating adversarial examples for each pruned network, we devise an adversarial example generation method to improve the efficiency of robustness evaluation.[8]

Our contributions are summarized as follows:

1. We present a novel framework, CCSRP, that considers network robustness during the pruning process and automatically solves the three-objective robust pruning problem through cooperative co-evolution. To our knowledge, this is the first application of EAs to robust pruning of spiking neural networks.

2. We introduce an adversarial example generation method to enhance the efficiency of evaluating the robustness of pruned networks.

3. We compare CCSRP with previous methods through experiments on three network architectures and two datasets. Experimental results demonstrate that CCSRP can achieve performance comparable to state-of-the-art methods.

## 2        Related work

### 2.1        Spiking Neural Network Pruning

The objective of pruning in spiking neural networks (SNNs) is to enhance operational efficiency by eliminating unnecessary components. Current pruning strategies are predominantly categorized into two types: unstructured and structured pruning.[9] Unstructured pruning involves direct adjustments to the weights within the network, theoretically offering significant computational speed-ups. However, the resulting sparse matrices and discontinuous structures are often incompatible with existing software and hardware environments, making practical acceleration challenging to achieve. In contrast, structured pruning targets the systematic removal of components, such as filters in convolutional layers of SNNs, demonstrating superior performance in real-world applications, thereby gaining increased popularity and attention.[10]

Regarding the identification of redundant components, previous structured pruning approaches can be broadly divided into criteria-based and learning-based methods.[11] Criteria-based methods rely on expert-designed rules to identify and prune non-essential components, whereas learning-based methods utilize auxiliary modules to assess the importance of components for subsequent pruning. Nevertheless, both methods heavily depend on domain expertise, restricting their widespread application and flexibility.[12]

To reduce the dependency on expert knowledge, employing evolutionary algorithms (EAs) for the automatic discovery of optimized pruned network architectures emerges as a natural solution. Despite this, the vast search space of SNNs presents a significant challenge to EAs.[13] Recently, an innovative pruning method inspired by cooperative co-evolution, named CCEP, has been introduced. It adopts a divide-and-conquer strategy to tackle the immense search space, demonstrating impressive performance and underscoring the substantial potential of EA-based methods in the domain of neural network pruning. However, previous EA-based pruning approaches have not taken into account the robustness of the network, a factor of critical importance for many application scenarios.[14]

### 2.2        Robust Spiking Neural Network Pruning

Recent studies have explored the relationship between robustness and network capacity, uncovering that a sub-network of the original network may exhibit similar or even superior robustness compared to the original network, with considerable variance in robustness among different sub-networks. This discovery has spurred research into robust pruning for spiking neural networks (SNNs), aiming to identify a compact SNN that retains robustness.[15] The existing handful of methods typically employ adversarial training to train a network and proceed with unstructured pruning based on criteria designed by experts. For instance, ADV-LWM prunes weights with small l1-norms and fine-tunes the resulting network through adversarial training to regain robustness. Utilizing the ADMM pruning framework, one approach replaces the original training loss with an adversarial counterpart. HYDRA assigns importance scores to all weights within the network, optimizing the adversarial loss by adjusting these scores while freezing the weights. Subsequently, weights with minimal importance scores are pruned. DNR opts to prune filters corresponding to feature matrices with small Frobenius norms. Moreover, these methods necessitate appropriate pruning ratios for each layer, which often requires extensive expert knowledge and iterative experimentation.[16]

## 3     CCSRP Method

Consider $N$ to be a thoroughly trained neural network composed of $n$ convolution layers, designated as $L_1, L_2, ..., L_n$, where each layer $L_i$ contains $l_i$ filters and $L_{ij}$ refers to the jth filter of the ith layer. The process of robust pruning in neural networks is an optimization challenge aimed at isolating a group of filters within $N$. The goal is to enhance both the network's accuracy and its robustness against perturbations while concurrently reducing the computational burden. We define a mask vector $M = \{m_{ij}|m_{ij} \in \{0,1\}, i \in \{1,2,...,n\}, j \in \{1,2,...,l_i\}\}$ with $m_{ij} = 1$ signifying the retention of filter $L_{ij}$. Therefore, we can represent a pruned network with the mask $M$ as follows:

$$N_M = \bigcup_{i=1}^{n} \bigcup_{j=1}^{l_i} m_{ij} L_{ij}$$

The performance of the pruned network $N_M$ is measured by $ACC(N_M)$, indicating accuracy on standard datasets, and $ACC_r(N_M)$, indicating robustness as measured by accuracy against adversarial examples. Additionally, $FLOP_s(N_M)$ accounts for the floating-point operations count, which assesses the computational expenses. We pose the robust pruning task as:

$$argmax_M \big(ACC(N_M), ACC_r(N_M), - FLOPs(N_M)\big)$$

Given that a Spiking Neural Networks (SNNs) can have a substantial quantity of filters eligible for pruning, identified by $\sum_{i=1}^{n} l_i$, this scenario poses a substantial optimization conundrum. To address this, we introduce CCSRP, a cutting-edge method for robust pruning. Drawing on the concepts from CCEP and CCRP, we utilize a cooperative coevolution-based pruning framework. This stratagem segments the search space into individual layers and applies evolutionary algorithms (EAs) to each layer in isolation. We prioritize robust accuracy as the main optimization criterion, steering the pruning process of neural networks in the direction of robustness. It's noteworthy to mention that evaluating $ACC_r(N_M)$ can be labor-intensive, as it requires generating bespoke adversarial examples for each pruned network variant. To circumvent this issue, we suggest an adversarial example generation technique that obviates the need for producing such examples for each pruned network iteration.[17]

### 3.1     EA

Algorithm 1 delineates the procedure of the CCSRP framework. This framework refines a well-trained neural network through iterative pruning, ultimately yielding a suite of pruned networks that maintain robustness for selection. The iterative process is articulated as follows. Initially, the algorithm generates a mask $M$ tailored to the network designated for pruning. Subsequently, this mask $M$ is partitioned into n segments corresponding to the hierarchical layers of the network. Following this, a set of adversarial examples $Da$ is produced for assessing the efficacy of the pruned network. For each segmented group, an evolutionary algorithm (EA) is employed to orchestrate optimization, procuring $m_i'$, indicative of the pruning outcome for the ith layer. By amalgamating $m_i'$ for all n layers and applying them to the baseline network Nb, the pruned network N' is procured. Post-pruning, to recoup the network's accuracy and robust accuracy, the pruned network N' undergoes fine-tuning via adversarial training. The fine-tuned model is then established as the new baseline network Nb for subsequent pruning in the next iteration and is archived in $H$. [18]

After T iterations, the CCSRP framework ceases and returns the pruned networks stored in archive H. An illustrative depiction of the CCSRP framework is also exhibited in Figure 1. Within each group of the evolutionary algorithm (EA) process, we engage a quintessential evolutionary procedure that commences with the random generation of an initial subpopulation. New individuals are bred by applying reproductive operators, followed by the evaluation of fitness for each, and the selection of the fittest individuals to advance to the succeeding generation. Upon reaching the termination criterion, the EA selects an individual from the ultimate subpopulation, which epitomizes the pruned layer for that group.[19]

The EA process within each group is elucidated in detail. Initially, it commences by generating an initial subpopulation $P$ consisting of $d$ individuals. An individual $|m|_0$, with all bits set to 1, is incorporated into $P$ to foster conservative pruning strategies. The remaining $d$-$1$ individuals are generated using a modified bitwise mutation operator with a mutation rate $p_1$. During each EA generation, $d$ new progenies are engendered by randomly choosing $d$ individuals from the subpopulation with replacement, followed by the application of a bitwise mutation operator with a mutation rate $p_2$. In accordance with CCEP, we have adapted the standard bitwise mutation operator to mitigate overly aggressive pruning. Specifically, a ratio bound r constrains the number of filters to be pruned. [20]

**Algorithm 1.** Combined CCRP with EA Optimization

**Input:** A well trained SNNs N with n layers, maximum number T of iterations, training; set Dt, a randomly sampled part Ds of the training set, adversarial data set Da, population size d, mutation rate p1, p2, ratio bound r, maximum number G of generations
**Output:** A set of pruned networks with different sizes

1: **Let** H = ∅, i = 0;
2: **Set** base network Nb = N;
3: **while** i < T **do**
4:      Generate a mask M based on Nb and initialize it
          with all bits equal to 1;
5:      **for** each layer l from 1 **to** n **do**
6:          **Set** mi as the lth segment of M;
7:          **Let** j = 0, m0 = mi;
8:          **Initialize** a subpopulation P with m0 and d-1
              individuals generated from m0 by applying
              the bitwise mutation operator with p1 and r;
9:          **while** j < G **do**
10:             **Uniformly** randomly select d individuals
                  from P with replacement as the parent
                  individuals;
11:             **Generate** d offspring individuals by apply
                  ing the bit-wise mutation operator with
                  p2 and r on each parent individual;
12:             **Calculate** the ACC and ACCr of d offspring
                  individuals by using Ds and Da; the
                  FLOPs of d offspring indi viduals;
14:             **Set** Q as the union of P and d offspring
                  individuals;
15:             **Rank** the 2d individuals in Q in descending
                  order by (ACC+ACCr)/2;
16:             **Replace** the individuals in P with the top
                  d individuals in Q;
17:             j = j + 1;
18:         **end while**
19:         **Select** the rank one individual in P as m'l;
20:     **end for**
21:     **Construct** new network N' by combining the m'n;
22:     Fine-tune N' with Dt; Nb = N';
24:     H = H ∪ Nb; i = i + 1;
26: **end while**
27: **return** H

Fitness evaluation for an offspring individual, given that each corresponds to a single pruned layer, involves splicing this layer with the other layers from the base network **Nb** to form a complete network. This allows for the assessment of accuracy, robust accuracy, and FLOPs of the offspring individuals. Specifically, accuracy (ACC) is gauged on the clean dataset $D_s$, sampled randomly from the training set $D_t$; robust accuracy (ACCr) is evaluated on the adversarial dataset $D_a$; and FLOPs are calculated directly.[21] Following evaluation, the **d** offspring and the **d** individuals in the current subpopulation **P** are amalgamated into a collective **Q**. Considering the three objectives, ranking individuals presents a complexity. For ease, individuals in **Q** are ranked in descending order based on the average of ACC and ACCr. Should there be a tie in average values, the individual with fewer FLOPs is favored. Alternative techniques, such as non-dominated sorting, may be employed and will be explored in future work. After **G** generations of evolution, the top-ranked individual from the final subpopulation is selected as the pruned outcome for the respective group.[22]

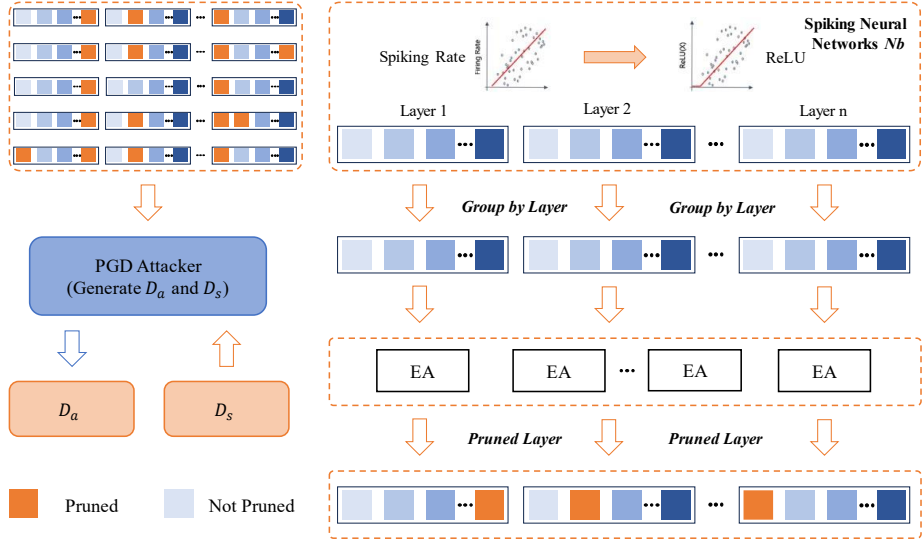

**Fig 1.** Illustration of the framework of CCSRP.

### 3.2 Robustness and Comparison

Typically, the robustness of spiking neural networks is gauged by their resilience to adversarial attacks. In this study, we define robust accuracy (ACCr) as the measure of robustness, determined by the network's performance against crafted adversarial examples.[23] We employ the advanced PGD white-box attack algorithm to generate these examples, noted for its iterative and time-intensive nature.[24] To circumvent the prohibitive computational expense of generating unique adversarial samples for each pruned network, we have devised an efficient generation method. This technique produces a shared adversarial dataset **Da** during a single iteration of CCSRP by apply-ing a mutation operation to select layers of the base network **Nb** and utilizing PGD. This procedure is independently replicated to compile **Da**, which assesses the robustness of all pruned networks during the current CCRP cycle.[25] Employing various subsets of **Nb** allows for a more accurate evaluation of a pruned network's robustness. In comparing CCSRP with CCRP, the latter expands upon spiking neural network pruning by factoring in robustness, emphasizing not only accuracy and compactness but also robustness as an optimization objective. CCSRP integrates an adversarial example generation method to minimize the costs of robustness evaluation and incorporates adversarial training in the fine-tuning stage to preserve the pruned network's robustness.[26]

---

**Algorithm 2:** Adversarial Generating

---

**Input:**
- Base network Nb with n layers, representing the original neural network before pruning.
- A randomly sampled part Ds of the training set Dt, which is a subset of the original dataset used for generating adversarial examples.
- The number k of sampled sub-nets, indicating how many times the generation process is to be repeated to produce diverse adversarial examples.

**Output:**
- Adversarial dataset Da, a collection of adversarial examples used for testing the robustness of the neural network.
1: **Initialize** the adversarial dataset Da as an empty set and the counter i as 0.
2: **Begin** a loop that will iterate k times, where k is the number of adversarial examples sets you wish to generate.
3: Within each iteration, randomly **select** ⌈n/k⌉ layers from the base network Nb.
4: **Apply** a mutation operation with parameters p1 and r to these selected layers to obtain a sub-net N'.
5: **Use** the Projected Gradient Descent (PGD) attack on this sub-net N' using the sampled dataset Ds to generate a set of adversarial examples A.
6: **Add** the newly generated adversarial examples A to the adversarial dataset Da.
7: **Increment** the counter i by 1.
8: **Repeat** the steps from 3 to 7 until i equals k.
9: Once all iterations are complete, **return** the Da.
**End of Algorithm.**

---

## 4    Experiments and Results

Our experimental investigation is conducted in three dimensions. Initially, we compare the performance of CCSRP (Cooperative Coevolutionary Spiking Neural Network Pruning) with the cutting-edge CCRP approach. Subsequently, we expand several prevalent unstructured and structured pruning methods to include robust pruning and juxtapose them with CCSRP. The third dimension involves conducting iterative experiments to evaluate the consistency of CCSRP and the visualization of the pruned network architectures. This evaluation employs two renowned image classification datasets: CIFAR-10 and SVHN, as well as three prototype neural networks adapted into their spiking versions: SVGG, SResNet, and SWRN. [27]

**Table 1.** Comparison ACC, ACCr, and inference speed with unstructured robust pruning.

| Dataset | Model | Method | Base ACC (%) | Base ACCr (%) | Speed (images/s) |
|---------|-------|--------|--------------|---------------|------------------|
| CIFAR-10 | VGG-16 | ADV-LWM | 82.52 | 51.91 | 2082.13 |
| | | ADV-ADMM | 78.36 | 47.07 | 2114.77 |
| | | HYDRA | 82.73 | 51.93 | 2077.57 |
| | | CCRP | 81.32 | 61.34 | 6842.39 |
| | | **CCSRP** | **82.94** | **62.41** | **5873.12** |
| | WRN-28-4 | ADV-LWM | 85.6 | 57.2 | 4142.74 |
| | | ADV-ADMM | 78.22 | 51.56 | 4375.58 |
| | | HYDRA | 85.6 | 57.2 | 4016.55 |
| | | CCRP | 85.91 | 53.42 | 4737.09 |
| | | **CCSRP** | **85.91** | **53.41** | **5823.12** |
| SVHN | VGG-16 | ADV-LWM | 90.5 | 53.5 | 2308.65 |
| | | ADV-ADMM | 89.35 | 54.61 | 2322.72 |
| | | HYDRA | 90.5 | 53.5 | 2334.29 |
| | | CCRP | 86.86 | 53.18 | 11124.56 |
| | | **CCSRP** | **87.14** | **54.44** | **5973.12** |
| | WRN-28-4 | ADV-LWM | 93.5 | 60.1 | 5259.51 |
| | | ADV-ADMM | 92.14 | 59.07 | 5482.91 |
| | | HYDRA | 93.5 | 60.1 | 5294.31 |
| | | CCRP | 90.07 | 57.47 | 6467.55 |
| | | **CCSRP** | **89.98** | **56.41** | **5898.11** |

Adhering to standard filter pruning protocols, CCSRP targets all convolutional layers in SVGG and the first convolutional layer in the residual blocks of SResNet and SWRN for pruning. The popular adversarial training method TRADES is applied during both the pre-training and fine-tuning phases. CCSRP's parameter settings are as follows: it operates for 16 iterations (T=16), with the evolutionary algorithm (EA) for each group having a population size d of 5, mutation rates p1 and p2 set at 0.05 and 0.1 respectively, a ratio bound r of 0.1, a maximum of 10 generations G, and Ds constructed by randomly sampling 10% of the training set. When generating adversarial instances, the algorithm samples 5 sub-networks, with k fixed at 5.[28]

**Table 2.** Comparison ACC, ACCr, and pruning ratio with structured robust pruning.

| Dataset | Model | Method | Base ACC (%) | Base ACCr (%) | ACCl (%) | ACCr (%) | FLOPs (%) |
|---------|-------|--------|--------------|---------------|----------|----------|-----------|
| CIFAR-10 | VGG-16 | L1 | 81.57 | 61.71 | 2.00 | 3.37 | 69.23 |
| | | HRank | 81.91 | 61.11 | 7.05 | 3.01 | 65.85 |
| | | CCSRP | 82.57 | 61.61 | 0.15 | 6.32 | 77.84 |
| | ResNet-56 | L1 | 80.31 | 48.95 | 2.47 | -4.62 | 68.53 |
| | | HRank | 80.31 | 48.95 | 0.13 | -2.22 | 50.02 |
| | | CCSRP | 81.32 | 48.85 | 0.24 | -8.34 | 72.32 |
| | WRN-28-4 | L1 | 85.91 | 53.61 | 2.00 | 3.37 | 69.23 |
| | | CCSRP | 86.11 | 53.71 | -0.89 | -8.26 | 66.98 |
| SVHN | VGG-16 | L1 | 86.86 | 53.18 | 2.17 | 4.35 | 85.88 |
| | | HRank | 86.06 | 54.53 | 0.40 | 5.03 | 65.85 |
| | | CCSRP | 87.16 | 53.43 | -0.96 | 2.64 | 80.54 |
| | ResNet-56 | L1 | 85.91 | 52.24 | -2.04 | -1.78 | 60.08 |
| | | HRank | 87.09 | 55.57 | -1.53 | 3.25 | 50.02 |
| | | CCSRP | 86.91 | 52.32 | -1.93 | -2.11 | 70.78 |
| | WRN-28-4 | L1 | 90.07 | 57.47 | 1.45 | 4.19 | 72.11 |
| | | CCSRP | 90.47 | 57.34 | -1.24 | 1.56 | 71.23 |

Under the adversarial training framework employing TRADES, a standard configuration was adopted. Stochastic Gradient Descent (SGD) was chosen as the optimizer, with an initial learning rate set at 0.1, which was adjusted through a cosine annealing strategy during the fine-tuning phase. The weight decay coefficient was set at 0.0001, with a momentum value of 0.9. Each cycle of the fine-tuning process consisted of 30 training epochs, with a batch size of 128. During the PGD adversarial attack phase, the adversarial training stage was configured with a perturbation budget, iteration steps, and per-step perturbation values of 8/255, 10, and 2/255, respectively; for evaluation and testing phases, these values were adjusted to 8/255, 40, and 2/255. A comparative analysis of the CCSRP Performance Relative to three cutting-edge unstructured robust pruning techniques—ADV-LWM, ADV-ADMM, and HYDRA—as well as two structured pruning techniques applied in robust pruning scenarios—L1 and HRank—was conducted, also considering the latest CCRP method. Results for HYDRA and ADV-LWM were obtained from their published models. All experiments were conducted using PyTorch and performed on an A100 GPU.[29]

In the comparison with unstructured robust pruning techniques, the performance of CCSRP was first compared against leading unstructured robust pruning methods based on accuracy degradation, robust accuracy degradation, and inference speed, as detailed in Table 1.[30] Considering that the decrease in FLOPs for unstructured models does not accurately reflect computational efficiency in practical applications, inference speed was used as a measure of computational cost. The inference speed was tested under a batch size of 128 for 100,000 32×32-pixel images. For CCSRP, results from the 10th iteration were presented in Table 1 for comparison.[31] In most cases, CCSRP achieved lower accuracy and robust accuracy degradation, along with faster inference speed. Despite HYDRA and ADV-LWM surpassing CCSRP in terms of robust accuracy degradation on the SVHN dataset, they experienced greater accuracy degradation and slower inference speeds.[32] In comparison with structured robust pruning techniques, to enable a more comprehensive analysis, two structured pruning methods, L1 and HRank, were extended to robust pruning scenarios through the incorporation of adversarial training in both pre-training and fine-tuning steps. For CCSRP, results from the 16th iteration were selected for comparison. The outcomes in Table 2 indicate that, compared to L1 and HRank, CCSRP consistently achieved better performance on at least two evaluation metrics.[33]

## 5      Conclusion

This paper introduces a novel approach for automatic robust pruning of spiking neural networks, named CCSRP, aimed at enhancing the stability and performance of spiking neural networks when faced with various disturbances and anomalies. Unlike conventional pruning methods that focus solely on improving network performance, CCSRP treats robust pruning as an optimization problem that integrates accuracy, sparsity, and robustness, and employs an innovative adaptive co-evolution framework to solve this problem. The essence of this method lies not only in pursuing network performance optimization but also in enhancing the network's resistance to external interferences, thereby maintaining good stability and accuracy under various challenging conditions. To our knowledge, this is the first time evolutionary algorithms (EAs) have been applied to the pruning problem of spiking neural networks. This interdisciplinary innovative application not only offers a new perspective and method for optimizing the robustness of spiking neural networks but also opens up new avenues for using evolutionary algorithms to solve complex optimization problems. Comparative experiments demonstrate that CCSRP exhibits comparable or even superior performance in some aspects to existing techniques, further validating the effectiveness and potential of this method.

Future work will focus on two main areas: one is to conduct more in-depth theoretical analysis to better understand the behavior and performance of CCSRP under different conditions, and the other is to explore and apply more advanced multi-objective optimization techniques to further enhance the performance of CCSRP. In particular, by introducing the latest multi-objective optimization algorithms and strategies, we aim to more finely balance the relationship between robustness, sparsity, and accuracy, thereby enhancing the overall performance and applicability of spiking neural networks in a broader range of application scenarios.